# Deep SCNN-based Real-time Object Detection for Self-driving Vehicles Using LiDAR Temporal Data

**Shibo Zhou[1], Ying Chen[2], Xiaohua Li[1]**, *Senior Member, IEEE*, **Arindam Sanyal[3]**, *Member, IEEE*

[1]Department of Electrical and Computer Engineering, Binghamton University, The State University of New York, Binghamton, NY 13902 USA
[2]Department of Management Science and Engineering, School of Management, Harbin Institute of Technology, Harbin 150000 China
[3]Department of Electrical Engineering, The State University of New York at Buffalo, Buffalo, NY 14260 USA
Corresponding author: Ying Chen (e-mail: yingchen@hit.edu.cn).

The authors appreciate the valuable suggestions of Prof. Wenfeng Zhao of Binghamton University in the revision process. This research was partially supported by National Natural Science Foundation of China (Grant No. 91846301)

**ABSTRACT** Real-time accurate detection of three-dimensional (3D) objects is a fundamental necessity for self-driving vehicles. Most existing computer vision approaches are based on convolutional neural networks (CNNs). Although the CNN-based approaches can achieve high detection accuracy, their high energy consumption is a severe drawback. To resolve this problem, novel energy efficient approaches should be explored. Spiking neural network (SNN) is a promising candidate because it has orders-of-magnitude lower energy consumption than CNN. Unfortunately, the studying of SNN has been limited in small networks only. The application of SNN for large 3D object detection networks has remain largely open. In this paper, we integrate spiking convolutional neural network (SCNN) with temporal coding into the YOLOv2 architecture for real-time object detection. To take the advantage of spiking signals, we develop a novel data preprocessing layer that translates 3D point-cloud data into spike time data. We propose an analog circuit to implement the non-leaky integrate and fire neuron used in our SCNN, from which the energy consumption of each spike is estimated. Moreover, we present a method to calculate the network sparsity and the energy consumption of the overall network. Extensive experiments have been conducted based on the KITTI dataset, which show that the proposed network can reach competitive detection accuracy as existing approaches, yet with much lower average energy consumption. If implemented in dedicated hardware, our network could have a mean sparsity of 56.24% and extremely low total energy consumption of 0.247mJ only. Implemented in NVIDIA GTX 1080i GPU, we can achieve 35.7 fps frame rate, high enough for real-time object detection.

## I. INTRODUCTION

In recent years, increased attention has been paid to point cloud data processing for autonomous driving applications because of significant improvements in automotive light detection and ranging (LiDAR) sensors, which deliver three-dimensional (3D) point clouds of the environment in real time. Point cloud data have highly variant density distributions throughout the measurement area [14], which can be exploited for object detection [1, 7, 22]. Nevertheless, different from camera images, LiDAR point clouds are unordered and sparse, which results in some difficulties for real-time object detection.

To address the point cloud object detection challenge, many approaches have been proposed, which can be divided into three general classes. The first class project point clouds into a perspective view and detect objects via image-based algorithms [4, 12]. The second class convert point clouds into a 3D voxel grid and use hand-crafted features to encode each voxel [2, 22]. The third class are similar to the second class but change the hand-crafted features into machine-learned features [5].

Owing to the machine-learned features, the third-class can achieve much better object detection performance. Qi et al. [14] proposed the PointNet which learns point-wise features of point clouds using deep neural networks. Qi et al. [15] proposed PointNet++ to allow networks to learn local structures at different scales. Zhou and Tuzel [24] developed the VoxelNet method, which can learn discriminative feature representations from point clouds and predict accurate 3D bounding boxes in an end-to-end module. Simon et al. [19] developed Complex-YOLO, a real-time 3D object detector that uses an enhanced region-proposal network (E-RPN) to estimate the orientation of objects coded with imaginary and real parts for each box. Recently, Simon et al. [20] presented a novel fusion (i.e., Complexer-YOLO) of neural networks that uses a state-of-the-art 3D detector and visual semantic segmentation in the field of autonomous driving. The accuracy of these methods has been demonstrated with the KITTI vision benchmark dataset [3].

There is much less work that focuses on the energy consumption of real-time object detection, although low energy consumption is a critical requirement for many practical applications such as autonomous vehicles. Convolutional neural networks (CNNs) have been the most popular techniques for object detection [20, 24]. However, their high energy consumption has been a challenging issue. By comparison, it is well known that spiking neural networks (SNNs) are energy efficient and can potentially have orders-of-magnitude lower energy consumption than CNNs [21].

Although the investigation of SNNs is far less than CNNs, numerous studies have shown that SNNs are able to achieve similarly high image classification accuracy [8, 9]. One of the major challenges for SNNs lies in the non-differentiability of spiking activities which makes the training of large-scale complex networks difficult. Zhou et al. [23] proposed a direct training-based spiking-CNN (SCNN) that could recognize the CIFAR-10 dataset using much less energy than CNN while reaching the same state-of-the-art accuracy. Nevertheless, efficient SCNN methods have not yet been reported over more complex data sets such as the KITTI 3D point clouds.

For the point cloud data, a special issue is how to translate from point-cloud format into a spiking format suitable for SNNs. If the input data are camera images, the input image can be converted into spike trains based on pixel intensity [21] or encoded into spike times [9].

Unfortunately, this approach is not efficient enough for point cloud data because of the sparsity and non-even density distributions of point clouds.

In this paper, similar to [20] and [24], we first quantize the point cloud to a 3D voxel representation so as to reduce the input data amount. Then, we design an innovative data-preprocessing layer that converts the 3D voxels into spike signals by adding time information to each voxel. Using this special temporal coding method, the input data are converted into spike times directly, and this permits us to design SCNNs with energy-efficient temporal coding. Finally, we use such SCNNs to replace the CNNs of the YOLOv2 architecture [17] to develop a large-scale object detection network. Note that similar to [23], the network is an end-to-end object detection network that combines feature extraction and bounding-box prediction.

We evaluate our developed network on the bird's-eye view and 3D detection tasks provided by the KITTI benchmark. Implemented over a NVIDIA GTX 1080i graphical processing unit (GPU), experimental results show that our network can reach a high frame rate of 35.7 fps, enough for real-time operation. Additionally, the detection accuracies for cars, pedestrians, and cyclists can reach the state-of-the-art level. To show the potential energy efficiency of our network, we propose an analog circuit implementation of the spiking neuron, based on which our proposed network would consume an average of 0.247 mJ only for processing each frame. This connotes our proposed network's high performance and energy efficiency.

The contributions of this paper are listed as follows:
1. We develop a novel data preprocessing layer to add temporal information to voxel data. With such temporal coding, we develop SCNNs with sparse spiking patterns to save energy.
2. We combine the SCNNs with YOLOv2 architecture to develop an efficient object detection network. We propose two variants of the network: one with skip connection (SC) and one without SC. The network without SC is suitable for the current neuromorphic chips, while the one with SC can be implemented with future neuromorphic chips that support skip connections.
3. We provide an analog circuit to implement the non-leaky integrate and fire neuron used in our SCNNs, based on which the energy consumption of a spike is estimated. We also provide a way to estimate the sparsity of the network. Combining low spike energy consumption and high network sparsity, the overall network energy consumption will be much lower than existing models. Simulation results demonstrate the extremely low energy consumption of our network.

## II. NETWORK ARCHITECTURE

As shown in Fig. 1, the proposed network comprises three functional blocks: a point cloud data preprocessing layer (LiDAR spike generation), spiking-convolution layers (feature learning), and a detection layer. In the following subsections, we provide a detailed description of each of these three blocks.

### A. Preprocessing Layer

For the 3D point cloud data such as KITTI data, each point $i$ is described by a four-number vector $(x_i, y_i, z_i, r_i)$, where $x_i, y_i, z_i$ are the 3D position of the reflection object point, and $r_i$ is the received laser light reflection intensity. The LiDAR device emits a pulse, which is reflected from the object and received by the LiDAR reception device. If the laser emission equipment is the origin of the coordinate system, the distance between each reflection point $i$ and the laser emission equipment can be calculated as:

$$d_i = \sqrt{x_i^2 + y_i^2 + z_i^2}. \quad (1)$$

Considering the volume of the space that LiDAR scans, 3D point cloud data set is usually huge and sparse. Using the 3D point cloud directly as input to deep networks is not computationally efficient. To reduce complexity (or to compress the LiDAR data set), one of the ways is to quantize the point clouds with a 3D voxel representation [19]. We adopt this approach and consider the 3D region of $[0, 60]\text{m} \times [-40, 40]\text{m} \times [-2.73, 1.27]m$. Specifically, we consider all the KITTI data points with $x_i \in [0, 60]\text{m}$, $y_i \in [-40, 40]\text{m}$, and $z_i \in [-2.73, 1.27]\text{m}$. We quantize the region into $768 \times 1024 \times 21$ voxels with the size of each voxel cell approximately equals to $0.08 \times 0.08 \times 0.19$ m³, as shown in Fig. 2. The voxelated space is a regular 3D coordinate system $(x_v, y_v, z_v)$, with length $x_v \in [0, 767]$, width $y_v \in [0, 1023]$, and height $z_v \in [0, 20]$. With the voxel representation, we construct a 3D tensor with shape $768 \times 1024 \times 21$ from each KITTI 3D point cloud data file and use it as input to the deep networks.

One of the differences between our work and the existing voxel-based works is that we use propagation time as the value of each voxel (i.e., tensor element) rather than the received light intensity or the number of data points. As introduced in Behroozpour et al. [25], the round-trip delay of the LiDAR emitted light for each point can be calculated using the following equation:

$$t_i = \frac{2d_i}{c}, \quad (2)$$

where $c$ is the propagation speed of the laser pulse. Since $t_i$ is the time information of each point, it can be used to represent the spike time for SCNN object detection networks.

For the voxel $(x_v, y_v, z_v)$, the value $t_v$ is calculated according to $t_i$ of all points inside this voxel. Due to the sparsity of point cloud, there are a lot of voxels that have no data points. In this case, we let $t_v=0$. For the voxels that have one or more data points, we randomly select one data point and use its time information to represent the voxel, i.e., $t_v = t_i$ for a randomly selected data point $i$ in this voxel. Note that we tried using the average time $t_v = \frac{1}{I}\sum_{i=1}^{I} t_i$ for all the $I$ data points in this voxel, but found that this averaging method did not lead to obvious performance gain. Therefore, we adopt the simpler random selection method.

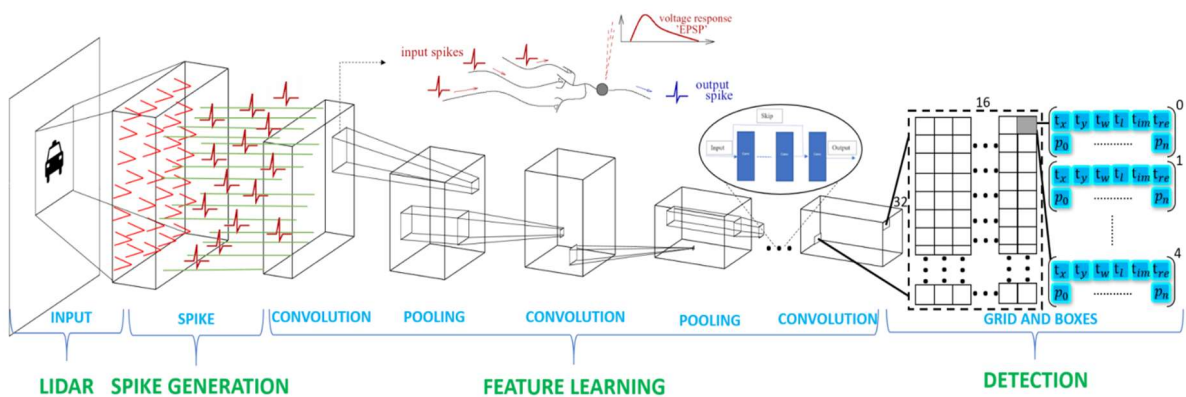

Figure 1. Proposed object detection architecture.

This propagation time-based encoding method is more desirable than other encoding methods. The intensity has large variations due to object shapes, materials, and environmental conditions such as raining, which increases the difficulty of deep network generalization. In contrast, the propagation time for each voxel is more regular. Compared with using the number of data points as voxel value, propagation time leads naturally to SCNN with a temporal coding that directly uses $t_v$ as spiking time. This temporal coding considers short-term stimuli to produce a small number of spikes [26], and thus leads to sparse spiking patterns.

After the LiDAR data to spike time conversion, the input to the first layer of the spiking-convolution is the 3D tensor with shape $768 \times 1024 \times 21$. The element value of the tensor is $t_v$. In the convolutional layers, we adopt 2D convolutions rather than 3D convolutions, where the filter sliding is conducted over the first two dimensions (length and width) only. For example, the first spike convolutional layer uses filters with shape $3 \times 3 \times 21$. In other words, we take the height dimension as the color dimension of the conventional images. The reason is that the height dimension has a shape 21, which is very small and much less than the other two dimensions. Using 2D convolution will lead to great reduction of the computational complexity but without obvious performance loss.

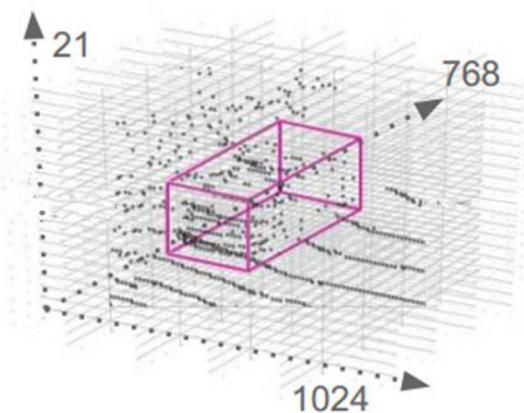

Figure 2. Voxelized point clouds within a 3D coordinate system.

### B. Spiking-Convolution Layers

Following Mostafa [9], Goltz et al. [28], Comsa et al. [29] and Kheradpisheh et al. [30], in the spiking-convolution layers, we use non-leaky integrate and fire neurons with current exponentially decaying synaptic kernels. A neuron's membrane dynamics are described by

$$\frac{dV_{mem}^j(t)}{dt} = \sum_i w_{ji} \sum_r k(t - t_i^r), \quad (3)$$

where the right-hand side of Eq. (3) is the synaptic current, $V_{mem}^j$ is the membrane potential of neuron $j$, $w_{ji}$ is the weight of the synaptic connection from neuron $i$ to neuron $j$, $t_i^r$ is the time of the $r^{th}$ spike from neuron $i$, and $k(x)$ is the synaptic current kernel given by

$$k(x) = \theta(x)exp(-\frac{x}{\tau_{syn}}), \text{where } \theta(x) = \begin{cases} 1 & \text{if } x \geq 0 \\ 0 & \text{otherwise} \end{cases} \quad (4)$$

The synaptic current jumps immediately when an input spike arrives. Then it decays exponentially with a time constant $\tau_{syn}$. Both $r$ and $\tau_{syn}$ are set to 1 for the rest of this paper.

We assume that a neuron receives $N$ spikes at times $\{t_1, \ldots, t_N\}$ with weights $\{w_1, \ldots, w_N\}$ from $N$ source neurons, and these spike times accumulate. As shown in Fig. 3, the neuron spikes when its membrane potential is over the firing threshold. After a spike, the membrane potential automatically resets to 0.

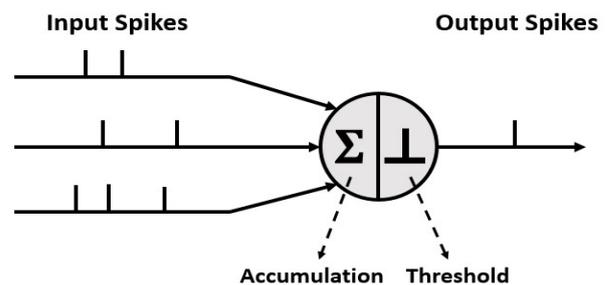

Figure 3. A model of spike neuron involving accumulating and thresholding operations.

If the neuron spikes at time $t_{out}$, the membrane potential for $t < t_{out}$ can be derived as

$$V_{mem}(t) = \sum_{i=1}^{N} \theta(t - t_i) w_i (1 - exp(-(t - t_i))). \quad (5)$$

Assume the thresholding membrane potential be 1. Then $t_{out}$ satisfies

$$1 = \sum_{i \in C} w_i (1 - \exp(-(t_{out} - t_i))), \quad (6)$$

where $C = \{i: t_i < t_{out}\}$. Therefore,

$$exp(t_{out}) = \frac{\sum_{i \in C} w_i exp(t_i)}{\sum_{i \in C} w_i - 1}. \quad (7)$$

We use this non-leaky integrate-and-fire neuron model to implement the convolutional neural networks. For example, in the first convolutional layer, the input neuron time ($t_i$ of Eq. (7)) is the voxel time $t_v$. The convolution kernel consists of weights $w_i$ in Eq. (7). The output neuron value is $t_{out}$, which is calculated by Eq. (7). As a result, communication between convolution layers occurs entirely through the temporal coded spike signals.

The convolution operation is illustrated in Fig. 4. Prior to matrix multiplication, we need to sort the spiking-time values $t_i$ in the small square of the input data in layer $n$, from small to large, into a vector $T$. The elements $w_i$ in the convolutional kernel are reordered to generate a matrix $W$ according to the changed position of spiking-time values. According to Eq. (7), for $w_i \in W$ and $t_i \in T$, a dot-product is performed between $w_i$ and $\exp(t_i)$. The neuron fires when the accumulated dot products reach the threshold. Consequently, the value of the mapping element, $t_{out}$, becomes the output. As noted, after the neuron fires, it is not allowed to fire again.

Two important marks are necessary to clarify. First, the sorting of $t_i$, which can be computationally demanding, is not needed when the SCNN is implemented in dedicated hardware. This is because $t_i$ represents the actual arrival time of the input neuron. Smaller $t_i$ means a spike arrives earlier and is thus accumulated earlier. Second, the input neurons with time $t_i > t_{out}$ do not participate in dot-product and accumulation. As a matter fact, these neurons would never fire spikes when implemented in dedicated hardware. Therefore, with temporal coding, the fired spikes can be quite sparse, which greatly conserves energy.

Using the SCNN to replace the CNN of the YOLOv2 backbone in [17], we obtain a new YOLOv2 network that detects and locates objects from spiking-time data. Combined with the real-time nature of YOLOv2, the presented network can be implemented efficiently on a neuromorphic architecture.

## C. Detection Layer

Following Simon et al. [19], we use E-RPN to derive the object's position $\{b_x, b_y\}$, length $b_l$, width $b_w$, probability $p_0$, class scores $\{p_1, …, p_n\}$, and the orientation $b_\emptyset$. To achieve proper orientation, we exploit the updated Grid-RPN approach from [19] to obtain

$$b_x = \sigma(t_x) + c_x, \quad (8)$$
$$b_y = \sigma(t_y) + c_y, \quad (9)$$
$$b_w = p_w e^{t_w}, \quad (10)$$
$$b_l = p_l e^{t_l}, \quad (11)$$
$$b_\emptyset = arg(|z|e^{ib_\emptyset}) = \arctan_2(t_{Im}, t_{Re}), \quad (12)$$

where $t_x$, $t_y$, $t_w$, $t_l$, $t_{Im}$ and $t_{Re}$ are 6 coordinate parameters for each bounding box that the network predicts. $(c_x, c_y)$ is the cell offset from the top left corner of the image. The bounding box prior has width and length $p_w$, $p_l$, respectively. In addition, $t_x$, $t_y$, $t_w$, $t_l$, $t_{Im}$ and $t_{Re}$ are the responsible regression parameters. With $t_x$, $t_y$, $t_w$, $t_l$ and $\arctan_2(t_{Im}, t_{Re})$, we can easily calculate the position, width, length and angle of each bounding box.

Our regression parameters are directly linked to the loss function $L_{loss}$ based on the Complex-YOLO of [19]. Specifically, the loss function is defined as

$$L_{loss} = L_{YOLO} + L_{Euler}, \quad (13)$$

where $L_{YOLO}$ is the sum of squared errors using the introduced multi-part loss, as shown in [17]. Additionally, according to [19], the Euler regression

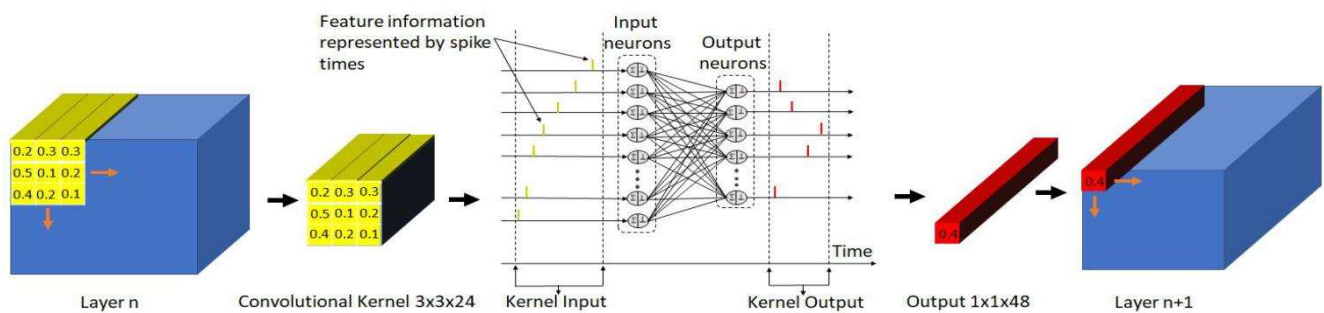

Figure 4. Illustration of spiking convolution in layer n.

part of the loss function, $L_{Euler}$, is defined as

$$L_{Euler} = \lambda_{coord} \sum_{i=0}^{s^2} \sum_{j=0}^{B} \alpha_{ij}^{obj} [(t_{im} - \hat{t}_{im})^2 + (t_{re} - \hat{t}_{re})^2], \quad (14)$$

where $\alpha_{sb}^{obj}$ indicates the $b^{th}$ bounding-box predictor in cell $s$, which has the highest Intersection over Union (IoU) in comparison with the ground truth for that prediction. $\lambda_{coord}$ is a scaling factor used to guarantee stable convergence during early phases. $\hat{t}_{im}$ and $\hat{t}_{re}$ are the estimated responsible regression parameters.

## III. ENERGY CONSUMPTION

Similar to [19] and [24], our network is a single-stage detector that can be trained in an end-to-end manner. The spike-based time coding strategy we use in this paper is similar to the scheme of time-to-first-spike in [13]. Hence, our proposed network will have lower running time than CNN-based models [5, 18]. Moreover, our network is more energy efficient because the network signals are transmitted via spikes. In this section, we analyze the energy consumption by first proposing an analog circuit of the spiking neuron, and then pointing out the sparsity of our proposed network.

### A. An Energy-Efficient Analog Neuron Circuit

To estimate the potential energy consumption of each spike, we propose an analog circuit as shown in Fig. 5 to implement the non-leaky integrate and fire neuron formulated in Eq. (3) and (4).

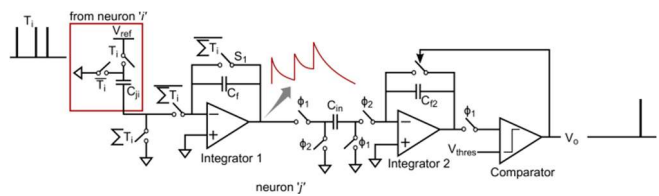

Figure 5. The analog circuit of the non-leaky integrate and fire neuron.

Fig. 5 shows the transistor-level schematic of a single neuron (neuron $j$). This neuron circuit comprises two analog switched-capacitor integrators and a comparator. The first integrator performs the summation operation in the right-hand-side of Eq. (3). When the neuron $i$ spikes, a capacitor $C_{ji}$ is charged to a fixed-potential $V_{ref}$, where the ratio of capacitor $C_{ji}/C_f$ represents weight of the synaptic connection from neuron $i$ to neuron $j$ ($w_{ji}$). The charge across $C_{ji}$ is dumped onto the feedback capacitor $C_f$ when the neuron $i$ is not spiking and the capacitor $C_f$ starts discharging through the switch connected parallel to it (denoted by $S_1$ in Fig. 5). The second integrator samples the output of the first integrator and starts integrating it and stores the charge on the feedback capacitor $C_{f2}$. Once the output of the second integrator exceeds a threshold voltage, $V_{thres}$, a comparator fires and produces an output of '1' which represents firing of neuron $j$. The comparator output also resets the second integrator.

The amplifiers in the integrators are designed using simple two-stage operational-amplifier structure with 55dB gain. A strongARM latch architecture is used to design the comparator. Designed in 65nm CMOS process, the neuron consumes 19 pJ energy for 1MHz input pulses. Since each neuron is only allowed to spike no more than once, each spike is thus averaged to consume $p = 19$ pJ on the analog circuit. Static power consumed by the amplifiers dominates the energy consumption of each neuron. However, gain-bandwidth of the amplifiers scale with frequency of input spikes, and energy consumption of each neuron can be reduced for systems with low frequency of input spikes.

Since all the computations are done in analog domain, the proposed neuron does not need digital memory to store intermediate results [10, 27], and hence, eliminates frequent data movements between memory and computation circuits. Thus, the proposed analog neuron is expected to have significantly lower energy consumption compared to digital implementation of the neuron. The weights of the spiking neural network, i.e, strength of synaptic connections between neurons, are encoded as capacitance values which further reduces storage requirement.

The implementation shown in Fig. 5 can be used when the SCNN has been trained and the weights are fixed. To make the neural weights programmable, we can apply non-volatile memory (NVM) on chip to make the capacitors programmable. For each capacitor, we can split it into several slices and the weights stored in the NVM can select the number of slices for each capacitor and set its weight. For a given application, if the weights are updated only once at start-up then the energy consumption due to memory access is not an issue. Memory access energy only becomes problematic if the memory has to be accessed every cycle of operation. Therefore, in a programmable deep network, the major cost would be the area increase since the memory has to be stored on the chip. The power increase is not going to be significant.

## B. Network Sparsity for Energy Efficiency

As seen in Eq. (6), the neuron spikes only when the membrane potential of the neuron reaches the threshold. After the neuron spikes, it is not allowed to spike again. To recognize the pattern or detect a scene, if the first $N$ neurons from the input layer can cause a neuron of the next layer to spike, the rest of the $M$ neurons in the input layer will not spike, which means a lot of neurons will not spike in each layer. Hence, we can calculate the sparsity of each layer using the following equation:

$$S_i = \frac{N_i}{M_i + N_i}, \quad (15)$$

where $S_i$ is the sparsity of the $i^{th}$ layer, $N_i$ is the number of spiking active neurons in the $i^{th}$ layer, and $M_i$ is the number of non-active neurons in the $i^{th}$ layer. The overall sparsity of a $K$-layer network is

$$S_{total} = \frac{N_{total}}{M_{total} + N_{total}} = \frac{\sum_{i=1}^{K} N_i}{\sum_{i=1}^{K} M_i + \sum_{i=1}^{K} N_i}, \quad (16)$$

where $M_{total}$ is the total number of non-active neurons, and $N_{total}$ is the total number of active neurons. The total energy consumption of the network can be calculated as

$$P = N_{total} \times p, \quad (17)$$

where $p$ is the energy consumed by each spike.

## IV. TRAINING AND EXPERIMENTS

In this section, we describe the proposed network and its training for the evaluation of the KITTI data in details.

### A. Training Details

We evaluated the proposed network using the challenging KITTI object detection benchmark, which contained 7481 and 7518 samples for training and testing, respectively. Because the input was LiDAR data, we only focused on birds-eye-view and 3D object detection for cars, pedestrians, and cyclists. Similar to the literature, each class was evaluated based on three difficulty levels (i.e., easy, moderate, and hard), considering the object's size, distance, occlusion, and truncation.

The detailed architecture of the proposed network is given in Table 1. Although our proposed network was based on the Complex-YOLO design from [19], we used only 9 spiking-convolutional layers, 1 traditional convolutional layer, 5 maxpool layers, and 3 intermediate layers. As a comparison, [19] used 18 convolutional layers, 5 maxpool layers, and 3 intermediate layers. Our proposed network was comparatively simpler than that of [19]. From the perspective of energy consumption, the simpler the network architecture, the less energy the network will consume.

As a special note, in the last layer, we used a traditional convolutional layer instead of a spiking-convolutional layer. Using the spiking-convolutional layer as the last layer would degrade performance compared with the current ones, which was a surprising observation we obtained from our preliminary test. A heuristic explanation is that because for SCNN the input values were time information, negative values were not allowed. But the values of some coordinates of the real 3D LiDAR data were negative based on the range presented in Section II. Traditional convolutional layers could handle negative values with the linear activation function $f(x) = x$.

Another special note is that we tested our network both with skipped connection (SC) or without skipped connection. As introduced by He et al. [6] and Orhan and Pitkow [11], SCs are simply extra connections between nodes at different layers of a neural network that skip one or more layers of nonlinear processing. They can improve the training of very deep neural networks without changing their main structure. Hence, in our network, we used SCs to improve the performance. However, latest detection chips (e.g., HTPU from CORAL and Akida from BrainChip) do not support SCs. The main reason is that the nodes of the hardware do not support event-packet synchronization across multiple layers. An SC also consumes extra energy in practice. Due to these considerations, we also experimented our network without SC.

The network was trained from scratch using stochastic gradient descent with a weight decay of 5e−4 and momentum of 0.9. The implementation was based on a modified version of the YOLOv2 framework [17]. Because the proposed model was a supervised learning-based network and the testing samples in the KITTI dataset had no labels, following the practice of [1, 19, 24], we divided the training set with an available ground truth and allocated 85% data for training and 15% for testing. To train the model well, during the first epoch, we started with a learning rate at 5e−5 to ensure convergence. After four epochs, we scaled the learning rate up to 5e−4 and gradually decreased it up to 1000 epochs.

Table 1. Proposed network architecture

| Layer | filters | size | input | output |
|---|---|---|---|---|
| Spike-conv | 32 | 3x3/1 | 768×1024×21 | 768×1024×32 |
| max | | 2x2/2 | 768×1024×32 | 384×512×32 |
| Spike-conv | 48 | 3x3/1 | 384×512×32 | 384×512×48 |
| max | | 2x2/2 | 384×512×48 | 192×256×48 |
| Spike-conv | 64 | 3x3/1 | 192×256×48 | 192×256×64 |
| max | | 2x2/2 | 192×256×64 | 96×128×64 |
| Spike-conv | 128 | 3x3/1 | 96×128×64 | 96×128×128 |
| max | | 2x2/2 | 96×128×128 | 48×64×128 |
| Spike-conv | 256 | 3x3/1 | 48×64×128 | 48×64×256 |
| Spike-conv | 1024 | 3x3/1 | 48×64×256 | 48×64×1024 |
| Spike-conv | 512 | 3x3/1 | 48×64×1024 | 48×64×512 |
| max | | 2x2/2 | 48×64×512 | 24×32×512 |
| Spike-conv | 1024 | 3x3/1 | 24×32×512 | 24×32×1024 |
| route | 9 | | | |
| reorg | | /2 | 48×64×256 | 24×32×1024 |
| route | 13 | | | |
| Spike-conv | 1024 | 3x3/1 | 24×32×2048 | 24×32×1024 |
| conv | 75 | 1x1/1 | 24×32×1024 | 24×32×75 |
| loss | | | **24×32×5×15** | |

Based on Eq. (15) and (16), we obtained the sparsity of each layer and the sparsity of the overall network for each sample. There were 1122 samples for validation, and the minimum, maximum, and mean sparsity values of our networks are given in Table 2. For the two trained networks (with SC, and without SC), there was no major difference between their maximum and minimum sparsity. The mean sparsity of the networks was 56.24% for the KITTI dataset.

Assume each spike consume 19 pJ. The energy consumption of the network was thus 0.247 mJ because the number of active neurons was about 13 million on average. Note that the estimated energy consumption does not include the energy spent in the last layer of the network and the preprocessing layer because these two layers do not involve spiking neurons. In CNN-based networks and many other SNN-based networks, all neurons are used for object detection or recognition. Thus, their energy consumption should be much higher than our obtained one. The small energy consumption value connotes the energy efficiency of our proposed network.

Table 2. Sparsity of the network for KITTI dataset.

| | Minimum | Maximum | Mean |
|---|---|---|---|
| Sparsity | 54.08% | 58.41% | 56.24% |

**B. Experiments**

We set up our experiments following the official KITTI evaluation protocol, where the IoU thresholds were 0.7 for the Car class, and 0.5 for Pedestrian and Cyclist classes. The IoU threshold was the same for both the bird's-eye view and full 3D evaluation. We compared the methods using the average precision (AP) metric.

**Evaluation in the Bird's- Eye View**

Our evaluation results for bird's-eye view detection are given in Table 3. Simon et al. [19] compared their proposed model, Complex-YOLO, with the first five leading models presented in Table 3 and demonstrated that their model outperformed all five in terms of running time and efficiency. They were still able to achieve detection accuracy comparable with the state of the art. As noted, although Complexer-YOLO [20] was more complicated than Complex-YOLO, the detection accuracies for all classes were lower than that of Complex-YOLO. Hence, we first focused on the comparison between Complex-YOLO and our network. As seen, all accuracy values of our proposed network with SC for detecting the car, pedestrian, and cyclist were higher than those using Complex-YOLO. Our network with SC showed better performance in object detection than our network without SC. Therefore, in the sequel, we mainly consider the model with SC and compare it with the others.

Table 3. Performance comparison for birds-eye-view detection: APs (in %) for our proposed networks compared with existing leading models.

| Method | Data | FPS | Car | | | Pedestrian | | | Cyclist | | |
|---|---|---|---|---|---|---|---|---|---|---|---|
| | | | Easy | Mod. | Hard | Easy | Mod. | Hard | Easy | Mod. | Hard |
| MV3D [1] | LiDAR + Mono | 2.8 | 71.09 | 62.35 | 55.12 | - | - | - | - | - | - |
| F-PointNet [16] | LiDAR +Mono | 5.9 | 81.20 | 70.39 | 62.19 | **51.21** | 44.89 | 40.23 | 71.96 | 56.77 | 50.39 |
| AVOD [7] | LiDAR + Mono | 12.5 | 73.59 | 65.78 | 58.38 | 38.28 | 31.51 | 26.98 | 60.11 | 44.90 | 38.80 |
| AVOD-FPN [7] | LiDAR + Mono | 10.0 | 81.94 | 71.88 | 66.38 | 50.80 | 42.81 | 40.88 | 64.00 | 52.18 | 46.61 |
| VoxelNet [24] | LiDAR | 4.3 | 81.97 | 65.46 | 62.85 | 57.86 | **53.42** | 48.87 | 67.17 | 47.65 | 45.11 |
| Complex-YOLO [19] | LiDAR | 50.4 | 67.72 | 64.00 | 63.01 | 41.79 | 39.70 | 35.92 | 68.17 | 58.32 | 54.30 |
| PointRCNN [31] | LiDAR | 10 | **86.96** | **75.64** | **70.70** | 47.98 | 39.37 | 36.01 | **74.96** | 58.82 | 52.53 |
| MLOD [32] | LiDAR + Mono | 8.3 | 77.24 | 67.76 | 62.05 | 47.58 | 37.47 | 35.07 | 68.81 | 49.43 | 42.84 |
| SCNet [33] | LiDAR | 25 | 83.34 | 73.17 | 67.93 | 47.83 | 38.66 | 35.70 | 67.98 | 50.79 | 45.15 |
| Our network with CNN | LiDAR | **51.2** | 65.13 | 61.82 | 59.57 | 37.93 | 34.76 | 31.03 | 64.79 | 55.52 | 51.29 |
| Our network with SC | LiDAR | 35.7 | 71.76 | 67.43 | 65.63 | 47.07 | 42.53 | 39.36 | 69.16 | **59.24** | **55.25** |
| Our network without SC | LiDAR | 35.7 | 68.94 | 66.65 | 65.39 | 38.03 | 32.43 | 28.64 | 58.40 | 50.61 | 45.75 |

Running the experiments with a NVIDIA GTX 1080i GPU, our network achieved frame rate 35.7 fps. Although this frame rate was lower than the 50.4 fps of Complex-YOLO, it was still much higher than those of the other five models. We also compared our detection results with the latest ones in [31-33]. Even though Point-RCNN [31] in the Car and Cyclist (easy and moderate) detection had higher accuracies than the others, its frame rate was low. As observed in Table 3, at the hard cyclist level, our proposed model had higher accuracies than the others. Its accuracies on other levels were competitive to the others.

As an ablation study, following the architecture in Fig. 1, we used CNN instead of SCNN to build the network. The results of this CNN-based network had higher frame rate but had much lower accuracy in all levels than the SCNN network, which indicates the benefits of using SCNN.

**3D Object Detection**

Apart from the bird's-eye view detection, we applied our proposed network to 3D object detection. The results are presented in Table 4. Similar to [19], we did not directly estimate the height information with regression but instead used a fixed spatial height location extracted from ground truth to implement the 3D object detection. As seen in Table 4, the detection accuracies of our network with SC on the car, pedestrian, and cyclist were all better than those of Complex-YOLO. Additionally, the accuracies of the proposed network with SC were comparable to the other models. Moreover, our network reached its highest accuracy with the moderate and hard cyclist levels. The proposed network with SC in all cases showed better performance than the network without SC and the ablation network with CNN only.

To illustrate the detection performance of our proposed network with SC, several 3D detection examples are presented in Fig. 6. For better visualization, we projected 3D boxes detected using LiDAR onto the red–green–blue (RGB) images. As seen in Fig. 6, the proposed network with SC had highly accurate 3D bounding boxes in all categories.

**V. CONCLUSION**

Existing LiDAR-based 3D real-time object detection methods use CNN. Although they can achieve high detection accuracy, their high energy consumption is a great concern for practical vehicular applications. This paper is the first to report the development of an SCNN-based YOLOv2 architecture for real-time object detection over the KITTI 3D point-cloud dataset considering the energy consumption. We designed a novel data preprocessing layer to translate the 3D point clouds directly into spike times. To better show the energy efficiency of the proposed network in real-time object detection, we built an analog neuron circuit to obtain the energy cost of each spike. We also proposed an energy consumption and network sparsity estimation method. Our proposed network had a mean spiking sparsity of 56.24% and consumed an average of 0.247 mJ only, indicating higher energy efficiency. Experimental results over the KITTI dataset demonstrated that our proposed network reached the-state-of-the-art accuracy in the bird's-eye view and full 3D detection. In some cases, our proposed network performed better than other typical models reported in literature.

Table 4. Performance comparison for 3D object detection: APs (in %) for our proposed networks compared with existing leading models.

| Method | Modality | FPS | Car | | | Pedestrian | | | Cyclist | | |
|---|---|---|---|---|---|---|---|---|---|---|---|
| | | | Easy | Mod. | Hard | Easy | Mod. | Hard | Easy | Mod. | Hard |
| MV3D [1] | LiDAR + Mono | 2.8 | 86.02 | 76.90 | 68.49 | - | - | - | - | - | - |
| F-PointNet [16] | LiDAR +Mono | 5.9 | 88.70 | 84.00 | 75.33 | **58.09** | **50.22** | **47.20** | 75.38 | 61.96 | 54.68 |
| AVOD [7] | LiDAR + Mono | 12.5 | 86.80 | 85.44 | 77.73 | 42.51 | 35.24 | 33.97 | 63.66 | 47.74 | 46.55 |
| AVOD-FPN [7] | LiDAR + Mono | 10.0 | 88.53 | 83.79 | 77.90 | 50.66 | 44.75 | 40.83 | 62.39 | 52.02 | 47.87 |
| VoxelNet [24] | LiDAR | 4.3 | 89.60 | 84.81 | 78.57 | 65.95 | 61.05 | 56.98 | 74.41 | 52.18 | 50.49 |
| Complex-YOLO [19] | LiDAR | 50.4 | 85.89 | 77.40 | 77.33 | 46.08 | 45.90 | 44.20 | 72.37 | 63.36 | 60.27 |
| Point-RCNN [31] | LiDAR | 10 | **92.13** | **87.39** | **82.72** | 54.77 | 46.13 | 42.84 | **82.56** | **67.24** | 60.28 |
| MLOD [32] | LiDAR + Mono | 8.3 | 90.25 | 82.68 | 77.97 | 55.09 | 45.40 | 41.42 | 73.03 | 55.06 | 48.21 |
| SCNet [33] | LiDAR | 25 | 90.07 | 86.48 | 81.30 | 56.87 | 46.73 | 42.74 | 73.73 | 56.39 | 49.99 |
| Our network with CNN | LiDAR | **51.2** | 83.48 | 72.41 | 72.39 | 43.81 | 42.91 | 41.12 | 68.39 | 59.15 | 55.98 |
| Our network with SC | LiDAR | 35.7 | 86.54 | 81.97 | 79.11 | 50.27 | 48.21 | 46.47 | 75.57 | 65.48 | **63.21** |
| Our network without SC | LiDAR | 35.7 | 86.93 | 82.70 | 82.23 | 38.34 | 36.43 | 35.52 | 62.40 | 55.97 | 55.34 |

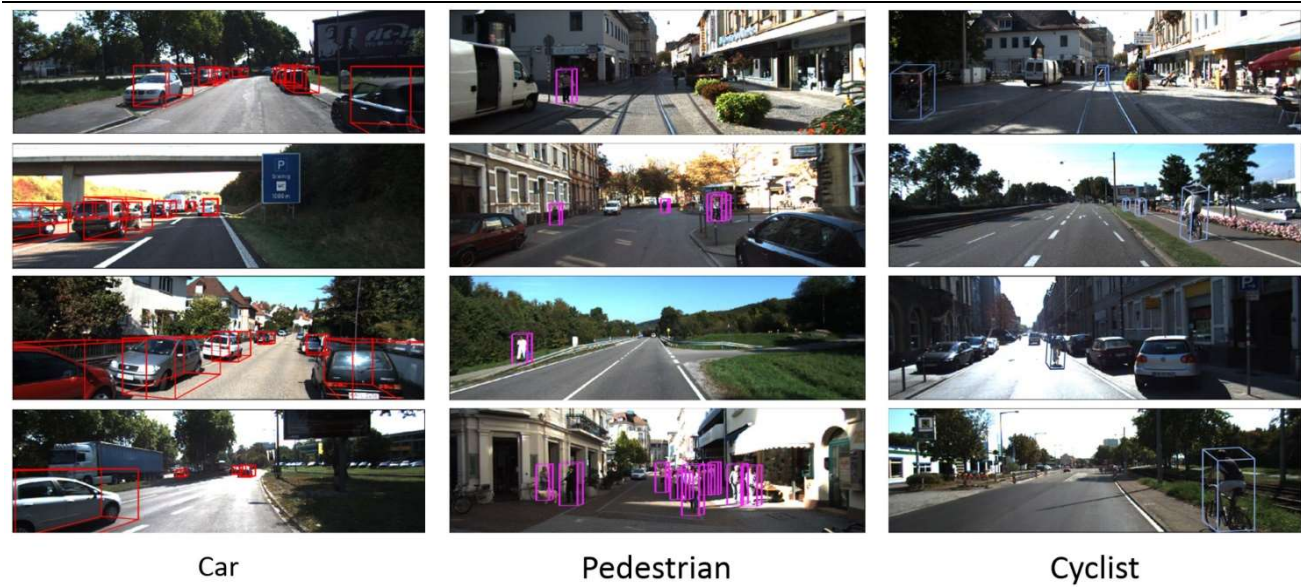

Figure 6. Qualitative results. 3D boxes detected with LiDAR are projected onto the RGB images.

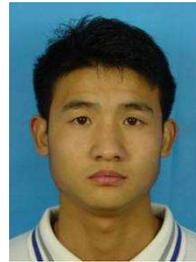

Shibo Zhou received the B.S. degree in Mechatronic Engineering from Hebei University of Architecture, Hebei, China, in 2013 and the M.S. degree in Mechanical Engineering from University of Texas at Arlington, Arlington, TX, USA, in 2016. He is currently pursuing the Ph.D. degree in Electrical Engineering at Binghamton University State University of New York, Binghamton, NY, USA. His research interest includes the low-power neuromorphic ASIC simulation and development, novel unsupervised learning algorithm, spiking neural network for object recognition and detection, computer vision and machine learning, deep neural network and 3D perception of self-driving car.

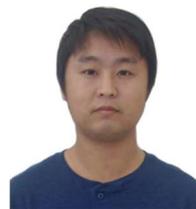

Ying Chen received the Ph.D. degree in Industrial Engineering from University of Texas at Arlington. He is currently an Assistant Professor in the School of Economics and Management at Harbin Institute of Technology. His research interests include data mining, machine learning, decision-making under uncertainty and optimization.

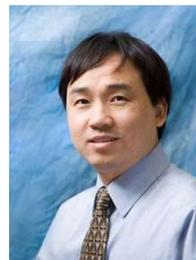

Xiaohua Li (M'00, SM'06) received the B.S. and M.S. degrees from Shanghai Jiao Tong University, Shanghai, China, in 1992 and 1995, respectively. He received the Ph.D. degree in Electrical Engineering from the University of Cincinnati, Cincinnati, OH, in 2000. He was an assistant professor from 2000 to 2006, and has been an Associate Professor since 2006, both with the Department of Electrical and Computer Engineering, State University of New York at Binghamton, Binghamton, NY. His research interests are in the fields of signal processing, machine learning, deep learning, wireless communications, and wireless information assurance.

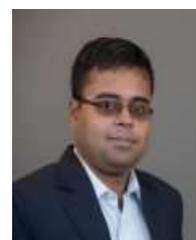

Arindam Sanyal (M'14) received his Ph.D. from The University of Texas at Austin in 2016, his M.Tech from The Indian Institute of Technology, Kharagpur in 2009 and B.E from Jadavpur University, India in 2007. He is an Assistant Professor in the Electrical Engineering Department at The State University of New York at Buffalo. Prior to this, he was a Design Engineer working on low jitter PLLs at Silicon Laboratories, Austin. His research interests include analog/mixed signal design, bio-medical sensor design, analog security and on-chip artificial neural network. He is the recipient of National Science Foundation CISE Research Initiation Initiative (CRII) award in 2020, Intel/Texas Instruments/Catalyst Foundation CICC Student Scholarship Award in 2014 and Mamraj Agarwal Award in 2001.